\documentclass[letterpaper, 10pt, conference]{ieeeconf}

\usepackage{lmodern}
\usepackage{amssymb}
\usepackage{graphicx}
\usepackage{color}
\usepackage{ulem}

\definecolor{red}{rgb}{0.8,0,0}
\definecolor{green}{rgb}{0,0.8,0}
\definecolor{blue}{rgb}{0,0,0.8}
\definecolor{yellow}{rgb}{0.8,0.8,0}
\definecolor{cyan}{rgb}{0,0.8,0.8}
\definecolor{black}{rgb}{0,0,0}

\newcommand{\revise}[1]{\textcolor{black}{#1}}

\newcommand{\bm}[1]{\mbox{{\boldmath $#1$}}}

\IEEEoverridecommandlockouts
\title{
{\LARGE \bf
Experiments on Learning Based Industrial Bin-picking with Iterative Visual Recognition
}
}

\author{
Kensuke Harada, Weiwei Wan, Tokuo Tsuji, Kohei Kikuchi, Kazuyuki Nagata, and Hiromu Onda
\thanks{K.~Harada and W~Wan are with Graduate School of Engineering Science, Osaka University, Toyonaka, Japan
       {\tt\small harada@sys.es.osaka-u.ac.jp}}%
\thanks{K.~Harada, W.~Wan, K.~Nagata and H.~Onda are with Intelligent Systems Research Institute, 
	National Institute of Advanced Industrial Science and Technology (AIST), Tsukuba, Japan}
\thanks{T.~Tsuji is with the Faculty of Engineering, Kanazawa University, Kanazawa, Japan}
\thanks{K.~Kikuchi is with Toyota Motors Co., Ltd., 1 Toyota-cho, Toyota 471-8572, Japan}%
}

\begin{document}

\maketitle
\thispagestyle{empty}
\pagestyle{empty}

\begin{abstract}
This paper shows experimental results on learning based randomized bin-picking combined with iterative visual recognition. 
We use the random forest to predict whether or not a robot will successfully pick an object for given depth images of the pile 
\revise{taking the collision between a finger and a neighboring object into account}. 
For the discriminator to be accurate, we consider estimating objects' poses by merging multiple depth images of the pile captured from different points of view by using a depth sensor attached at the wrist. 
\revise{We show that,} even if a robot is predicted to fail in picking an object with a single depth image due to its large occluded area, it is finally predicted as success after merging multiple depth images. 
In addition, we show that the random forest can be trained with the small number of training data. 
\end{abstract}

\section{Introduction}

Randomized bin-picking refers to the problem of automatically picking an object from randomly stacked pile. 
If randomized bin-picking is introduced to a production process, 
we do not need any part-feeding machines or human workers to once arrange the objects to be picked by a robot. 
Although a number of researches have been done on randomized bin-picking for industrial parts such as 
(Turkey, 2011; Kristensen, 2001; Frydental, 1998; Hujazi, 1990; Ghita, 2003; 
Kirkgaard, 2006; Fuchs, 2010; Zuo, 2004; Domae, 2014; Dupuis, 2008; Harada, 2013; Harada, 2014), 
randomized bin-picking is still difficult due to the complex physical phenomena of contact among objects and fingers. 
To cope with this problem, learning based approach has been researched by some researchers such as 
(Harada, 2016a; Harada, 2016b, Bousmalis, 2017; Mahler, 2017). 
By using the learning based approach, it is expected that the complex physical phenomena can automatically be learned and that the bin-picking can be easily performed with high success rate. 
However, there have been some problems in the learning based bin-picking methods. 
Firstly, although the 2D/3D image including occlusion has been used to predict whether or not a robot can successfully pick an object, a finger may be inserted into the occluded area while picking an object. 
In this case, a robot may fail in picking an object due to unexpected contact between a finger and a neighboring object. 
Secondly, the number of training data is sometimes extremely large (Levine, 2016). 

\begin{figure}[t]
	\centering
	\includegraphics[width=5.5cm]{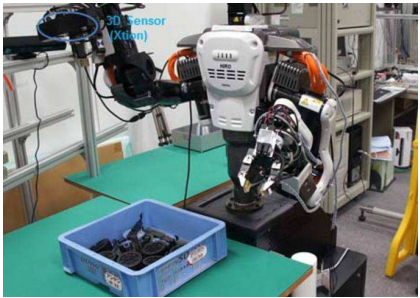}\\
	{\scriptsize (a) Overview of randomized bin-picking system}\\
	\includegraphics[width=4.5cm]{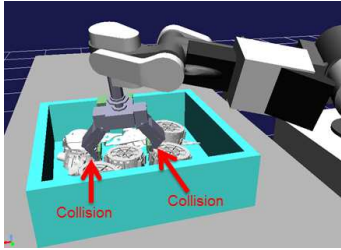}\\
        {\scriptsize (b) Failure of picking due to hand contact with neighboring objects}
	\caption{Overview of our bin-picking system \label{intro:RA-L2015}}
        \vspace{-5mm}
\end{figure}

We have proposed a learning based approach for randomized bin-picking (Harada, 2016a). 
\revise{The feature of our method is that, although randomized bin-picking often fails since a finger 
contacts a neighboring object which is not traversable, our learning method explicitly takes 
the contact between a finger and a neighboring object into account} (Harada, 2016a). 
In this method, we first detect the objects' poses from a 3D depth image (Fig. \ref{intro:RA-L2015}(a)). 
The depth image was also used to predict whether or not a robot can successfully pick one of the objects from the pile. 
Fig. \ref{intro:RA-L2015}(b) shows a scene in which a gripper tries to grasp one of the objects. 
Since objects are placed close to each other, a finger may contact a neighboring object while a gripper approaches the target object. 
In such cases, whether or not a robot can successfully pick an object from the pile depends on the configuration of neighboring objects, i.e., a robot will successfully pick an object if a finger contacts a neighboring object which is traversable. 

To predict whether or not a robot successfully picks an object from the pile, we need accurate 3D visual information near the target object. 
If the visual information includes a large occluded area, we may encounter the following two unexpected failure cases. 
Firstly, if the image of target object includes large occluded area, the pose \revise{of} the target object will be inaccurately identified. 
In this case, a robot may fail in picking an object since the grasping pose of the target object is different from its desired one. 
Secondly, if the neighboring objects are occluded, the prediction is made without taking these neighboring objects into consideration. 
In this case, a robot may fail in picking an object due to an unexpected contact between a finger and an occluded neighboring object. 
In our previous research (Harada, 2016a), 
occluded area included in a depth image prevented the prediction to be accurate. 

For the purpose of reducing the uncertainties in picking an object from the pile, we have proposed an iterative visual recognition method (Harada, 2016b) 
for randomized bin-picking. 
This method merges multiple visual images of the pile by capturing from different points of view by using a 3D depth sensor attached at the wrist. 
This method contributes to reducing the occluded area included in a depth image of the pile. 

On the other hand, the purpose of this research is to supply experimental results on learning based randomized bin-picking 
(Harada, 2016a) 
combined with iterative visual recognition (Harada, 2016b). 
We additionally impose penalty for candidates of grasping posture according to the occluded area. 
We observed cases where, even if a robot is predicted to fail in picking an object with a single depth image due to its occlusion, a robot is predicted to successfully picking an object after merging multiple depth images. 
In addition, we show that, if we construct a discriminator by using the random forest, the number of training data is much smaller than other existing methods. 

The rest of the paper is organized as follows. After introducing the related works in Section 2, we explain the learning based approach in Section 3. Section 4 explains a method for sensor pose calculation. 
Section 5 explains a method for object pose detection. 
We explain a method for picking task execution in Section 6. Finally, experimental results are \revise{shown} in Section 7. 

\section{Related Works}

The research on industrial bin-picking has been mainly done on image segmentation 
(Turkey, 2011; Kristensen, 2001; Frydental, 1998; Hujazi, 1990), 
pose identification 
(Ghita, 2003; Kirkgaard, 2006; Fuchs, 2010; Zuo, 2004), 
and picking method 
(Domae, 2014; Dupuis, 2008; Harada, 2013; Harada, 2014). 

As for the research on bin-picking method, 
Ghita and Whalan (Ghita, 2003) 
proposed to pick the topmost object of the pile. 
Domae et al. (Domae, 2014) 
proposed a method for determining the grasping pose of an object directly from the depth image of the pile. 
Some researchers such as (Fuchs, 2010; Dupuis, 2008; Harada, 2013; Harada, 2014) 
proposed a method for identifying the poses of multiple objects of the pile and picking one of them by using a grasp planning method. 
However, the performance of randomized bin-picking has been limited since it is difficult to model the complex physical phenomena of contact among the objects and fingers. 

On the other hand, learning based approach on randomized bin-picking is expected to break this barrier existing in the conventional randomized bin-picking (Levine, 2016; Harada, 2016a; Bousmalis, 2017; Mahler, 2017). 
Levine et al. (Levine, 2016) 
proposed a end-to-end approach by using deep neural network whose input is a 2D RGB image. 
However, they need the extremely large number of training data which was collected 800,000 times of picking trials for two months by using a 2D RGB image of the pile. 
On the other hand, recently there is a trial on reducing the effort to collect a number of training data by using 
a method so-called GraspGAN (Bousmalis, 2017) 
and a cloud database (Mahler, 2017). 
On the other hand, our learning based approach uses the random forest with the small number of training data. 
Our method also tries to obtain more accurate 3D depth image captured from different points of view used to predict whether or not a robot successfully picks an object from the pile. 

The learning approach has also been used for grasping a novel object placed on a table 
(Curtis, 2008; Lenz, 2015; Pas, 2015; Ekvall, 2007) 
and for warehouse automation (Zeng, 2017; Lin, 2017). 
Pas et al. (Pas, 2015) 
developed a method for learning an antipodal grasp of a novel object by using the SVM (support vector machine). 
Lenz et al. (Lenz, 2015) 
used deep learning to detect the appropriate grasping pose of an object. 
Zeng et al. (Zeng, 2017) 
proposed a learning based picking method used for warehouse automation. 
However, industrial bin-picking is different from the warehouse application since the grasped object \revise{does not exist} in our daily life and it is impossible to use the generalized object recognition methods. 

As for the research on obtaining maximum visibility, there have been several researches on obtaining the optimal camera position to get maximum visibility of an object such as 
(Olague, 2002; Sablatnig, 2003; Scott, 2003; Setiz, 2006; Wenhardt, 2006; Tarabanis, 1995; Harada, 2016b). 
However, these methods have not been applied for randomized bin-picking. 

On the other hand, this research considers performing experiments on learning based randomized 
bin-picking (Harada, 2016a) 
combined with iterative visual recognition realizing the maximum visibility of the pile (Harada, 2016b). 
\revise{Different from conventional approaches, our method explicitly considers the contact between a finger and a neighboring object. 
In this paper, we newly show several experimental results. }
We show that the number of training data of our method is much smaller than other existing methods. 
By combining the iterative visual recognition with learning based picking, we show cases 
\revise{
where, even} 
if a robot is predicted to fail in picking an object with a single depth image due to the existence of its occluded area, it is predicted to be a success after merging multiple depth images. 

\section{Learning Based Bin-Picking}

\label{sec:learn}

This section explains our learning based randomized bin-picking (Harada, 2016a). 
We especially explain how to construct the discriminator by using Random Forest predicting whether or not a robot successfully picks an object from the pile. 

\begin{figure}
	\centering
	\includegraphics[width=8cm]{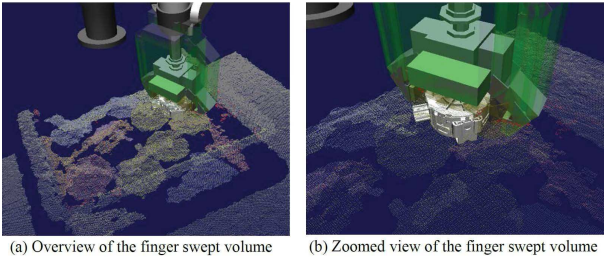}
	\caption{Swept volume of finger motion\label{fig:sweepvol}}
        \vspace{-0.3cm}
\end{figure}

Let us consider a two-fingered gripper to pick an object from the pile. 
To pick an object, a two-fingered gripper first moves from the approach pose to the preshaping pose (approaching phase), and then closes the fingers to grasp an object (grasping phase). 
As shown in Fig. \ref{fig:sweepvol}, we calculate the swept volume corresponding 
to the finger motion during the approaching and the grasping phases. 
Here, since the swept volume is used to see the point-cloud distribution of neighboring objects, it is calculated before the gripper actually moves. 
Given a point cloud of the pile and a grasping posture of a gripper, we consider removing the points belonging to the target object by checking the distance between a point and the surface of the target object. 
Then, we can obtain the distribution of point cloud of neighboring objects included in the swept volume of finger motion. 

\begin{figure}
	\centering
	\includegraphics[width=4.5cm]{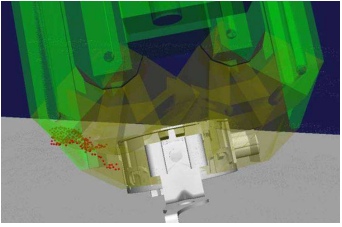}
	\caption{Point cloud included in the swept volume \label{fig:pointsVol}}
       \vspace{-0.4cm}
\end{figure}

\begin{figure}
	\centering
	\includegraphics[width=8cm]{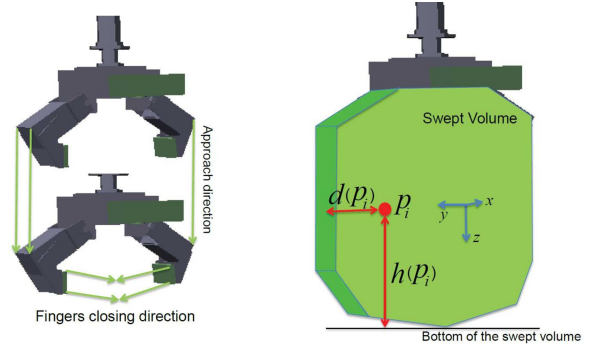}
	\caption{Definition of variables related to the finger swept volume \label{fig:heurist}}
\end{figure}

We construct a discriminator predicting whether or not a robot successfully picks an object based on the distribution of point cloud included in the swept volume of finger motion. 
For this purpose, we consider discretizing the area included in the swept volume and counting the number of points included in each discretized region. The number of points included in each discretized region defines a feature vector used in the random forest. 

In the following, we will describe more concretely how to formulate the feature vector. 
Let us introduce a coordinate system attached to the swept volume where $z$ and $x$ axes denote the approach direction and the direction perpendicular to the finger motion plane, respectively (Fig. \ref{fig:heurist}). 
$p_i$ $(i=1,\cdots,n)$ denotes the $i$-th point included in the swept volume. $d(p_i)$ denotes the minimum distance between $p_i$ and the boundary of the finger swept volume in the $y$-direction. 
$h(p_i)$ denotes the distance between $p_i$ and the bottom of the finger swept volume in the 
$z$-direction. 

Corresponding to the distance between a point and the boundary of the swept volume in the 
$y$ direction, we assume $b_y$ bins which width is $w_y$. 
Also, corresponding to the distance between a point and the bottom of the finger swept volume in the 
$z$ direction, we assume $b_z$ bins which width is $w_z$. 
The point $p_i$ $(i=1,\cdots,n)$ is stored to the $j_y$-th $(\le b_y)$ and the $j_z$-th $(\le b_z)$ bins in the $y$ and the $z$ directions, respectively, where their definitions are given by
\begin{eqnarray}
j_y &=& {\rm min}\left( \frac{d(p_i)}{w_y}, b_y \right), \\
j_z &=& {\rm min}\left( \frac{h(p_i)}{w_z}, b_z \right).
\end{eqnarray}

\begin{figure}
	\centering
	\includegraphics[width=7.5cm]{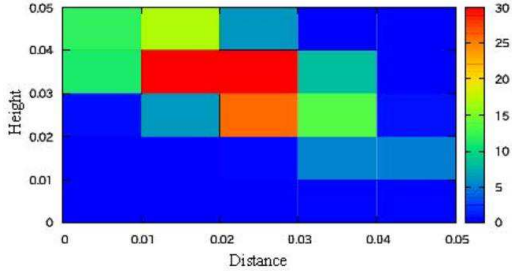}
	\caption{An example of the feature vector used in the random forest algorithm corresponding to the point-cloud distribution shown in Fig. \ref{fig:pointsVol} where the number of points of each bin is shown. \label{fig:forest}}
        \vspace{-0.4cm}
\end{figure}

\noindent
After capturing a point cloud used for the $j$-th picking trial, 
we count the number of points included in each bin for a give grasping configuration 
of the target object. 
Let $\bm{f}_{rj}$ be the $b_y b_z$ dimensional 
feature vector where each element is the number of points included in each bin. 
Fig.\ref{fig:forest} shows the feature vector corresponding to the grasping posture shown in 
Fig. \ref{fig:pointsVol} where we set $b_y=b_z=5$ and $w_y=w_z = 0.01${[m]}. 

Then, we explain how to formulate the Random Forest by using the obtained training data. 
By using the training data ${\cal L} = \left\{ (\bm{f}_{rj}, r_j), j=1,\cdots,m \right\}$ obtained through a series of bin-picking experiment, the Random Forest first generates $N$ subsets of training data denoted by 
${\cal L}_k$ $(k=1,\cdots,N)$ by randomly sampling the training data. For each subset, a decision tree is constructed. 
In case of the $k$-th decision tree $(k=1,\cdots,N)$, each node of the tree is a subset of ${\cal L}_k$. 
For example, let $\tilde{\cal L}_k$ be a subset of ${\cal L}_k$ forming a node of the $k$-th decision tree. 
To form its child nodes, we split $\tilde{\cal L}_k$ into $\tilde{\cal L}_k^L$ and $\tilde{\cal L}_k^R$ so as to minimize the Gini coefficient. We set the maximum depth of a decision tree to be $t_k$. 
From each decision tree, we can obtain the success rate of the pick. By the mean of the success rate obtained from all the decision trees, we finally estimate whether or not a robot successfully picks an object included in the pile. 

\section{Sensor Pose Calculation}

\label{sec:pose}

\revise{
For each picking trial, we first determine the sensor pose maximizing the visibility of the pile (Harada, 2016b). 
After taking the 3D depth image of the pile, we consider merging it to the previously taken one by using the method which will be 
explained in the next section. 
Then, we consider detecting the objects' poses as also explained in the next section. 
This sequence of operation is iterated as far as a robot continues to pick an object from the pile. 
This section especially explains how to determine the sensor pose. 
}

\begin{figure}[thb]
\centering
  \includegraphics[width=6cm]{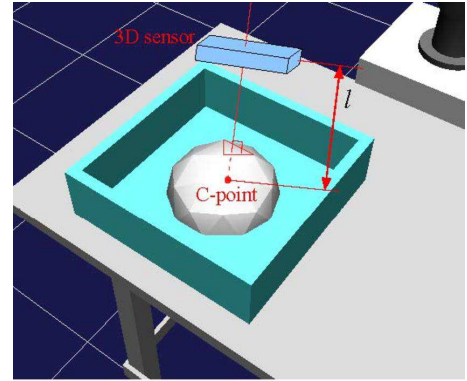}
  \caption{Regular polygon assumed at the geometrical center of bottom surface \label{fig:polygon}}
\end{figure}

\revise{
We first define a set of sensor pose candidates. 
Let us assume a $n$-faced regular polyhedron where its geometrical center is located at the center of the box's bottom surface 
(Fig. \ref{fig:polygon}). 
Let us also assume a line passing through the geometrical center and orthogonally intersecting a face of the polyhedron. 
To define sensor pose candidates, we assume a set of points along the line where the distance measured from the geometrical center is denoted by $l=l_1, l_2, \cdots, l_m$. 
At each point, we assume that the sensor faces the geometrical center. 
The following conditions are imposed for the sensor pose candidates. 
}

\revise{
\begin{itemize}
\item[] The sensor is located above the box's bottom surface. 
\item[] IK (inverse kinematics) is solvable. 
\item[] No collision occurs to the links. 
\end{itemize}
}

\begin{figure*}[thb]
\centering
  \includegraphics[width=12cm]{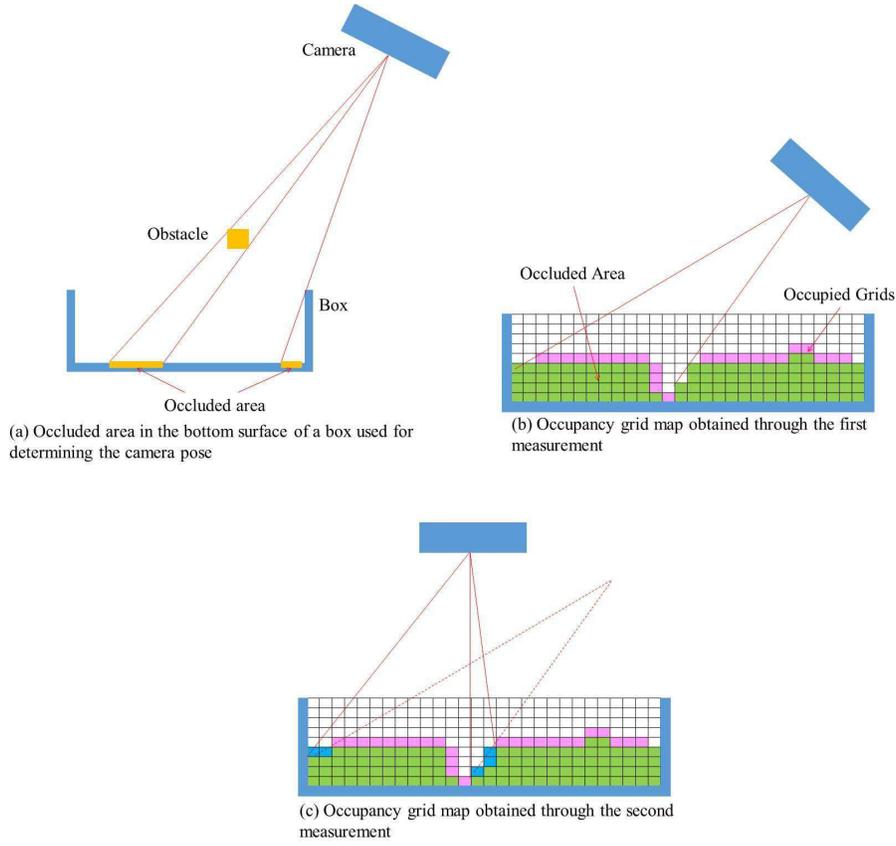}
  \caption{\revise{Determination of sensor pose maximizing the visibility of stacked objects where cells colord in pink and blue are the {\it occupied} and the {\it occluded} cells, respectively.} \label{fig:occupancy}}
\end{figure*}

\revise{
Among a set of candidates satisfying the above conditions, we consider selecting one maximizing the visibility of the pile. 
To define the visibility, we consider using the occupancy grid map by partitioning the storage area of a box into multiple grid cells (Thrun, 2005; Nagata, 2010). 
We mark {\it occupied} to the cells including the point cloud. 
For the first picking trial, we consider selecting a 3D sensor's pose minimizing the number of cells including the box's bottom surface (Fig. \ref{fig:occupancy} (a)). 
After the second picking trial, we consider using the depth image captured during the previous picking trial to determine the sensor pose as shown in Fig. \ref{fig:occupancy} (b) and (c). 
We mark {\it occluded} to the visible grid cells which are not marked as {\it occupied} in the previous picking trial. 
The sensor pose is determined to maximize the number of grid cells marked as {\it occluded}. 
Here, robotic bin-picking is usually iterated until there is no object remained in a box. 
This sequence of operation is iterated as far as a robot continues to pick an object from the pile. 
}

\section{Object Pose Detection}

\label{sec:est}

\begin{figure*}
	\centering
	\includegraphics[width=12.cm]{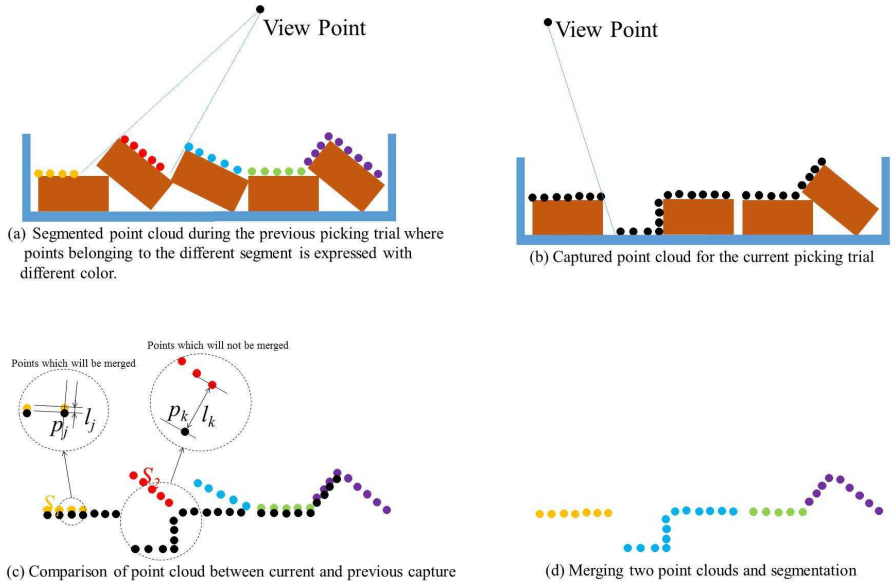}
	\caption{Segmentation of point cloud after the second picking trial \label{fig:merge}}
\end{figure*}

\revise{
This section explains a method for detecting the pose of randomly stacked objects. 
After capturing the depth image, we consider merging it to the previously captured image. 
Then, we consider segmenting the depth image (Stein, 2014) 
as shown in Fig. \ref{fig:merge}(a). 
For each segment whose bounding-box size is similar to that of an object, we try to estimate the pose of an object (Aldoma, 2012). 
}

\revise{
In the following, we mainly explain a method for merging the currently captured depth image to the previously captured one. 
The configuration of objects after a robot tries to pick an object is usually partially different from the configuration before the picking trial. 
We assume that the previously captured depth image has already been segmented (Stein, 2014). 
Fig. \ref{fig:merge} (a) shows the segmented point cloud obtained during the previous picking trial. 
On the other hand, Fig. \ref{fig:merge} (b) shows the current point cloud where the configuration of the objects is partially different from the previous one. 
If a segment of the previously captured depth image is similar to the current depth image, we consider merging the segment of previously depth image to the current one. 
By merging a part of the previous depth image captured from the different viewpoint to the current one, the occluded area is expected to be smaller. 
}

\revise{
The algorithm of merging the depth images is detailed in Fig. \ref{fig:merge} (c) and Algorithm 1. 
Let $\bar{P} = (\bar{p}_1, \bar{p}_2, \cdots, \bar{p}_m)$ and 
$P = (p_1, p_2, \cdots, p_n)$ be the previously captured point cloud and the current one, respectively. 
Also, let $\bar{P}_1, \bar{P}_2, \cdots$, and $\bar{P}_s$ be the segments of previous depth image. 
For each point included in the current depth image, we search for the point included in the previous depth image making the distance between them be minimum (lines 6 and 7). 
We then find a segment where the detected point belongs (line 8). 
For each segment of previous point cloud, we introduce two integer numbers ${\rm near}(i)$ and {\rm far}(i) expressing the number of points included in the segment $P_i$ where the minimum distance is smaller and larger, respectively, than the threshold {\rm MinDistance} (lines 9 and 10). We determine whether or not we merge the segment $\bar{P}_i$ into the point cloud $P$ depending on the ratio between ${\rm far}(i)$ and ${\rm near}(i)$. 
}

\bigskip

\noindent
{\bf Algorithm 1}\ \ Merging method between two point clouds\\
\\
1. for $i \leftarrow 1:s$ \\
2. \hspace*{0.5cm} near$(i)=0$ \\
3. \hspace*{0.5cm} far$(i)=0$ \\
4. end for\\
5. for $j \leftarrow 1:m$ \\
6. \hspace*{0.5cm} $d \leftarrow {\rm min}(|\bar{p}_1 - p_j| , \cdots, |\bar{p}_m-p_j|)$ \\
7. \hspace*{0.5cm} $k \leftarrow {\rm argmin}(|\bar{p}_1 - p_j| , \cdots, |\bar{p}_m-p_j|)$ \\
8. \hspace*{0.5cm} $t \leftarrow {\rm SegmentNumber}(\bar{p}_k)$ \\
9. \hspace*{0.5cm} if $ d < {\rm MinDistance}$ then : ${\rm near}(t) \leftarrow {\rm near}(t) + 1$ \\
10. \hspace*{0.43cm} else : ${\rm far}(t) \leftarrow {\rm far}(t) + 1$ \\
11. end for\\
12. for $i \leftarrow 1:s$\\
13. \hspace*{0.43cm} if $\frac{{\rm far}(i)}{{\rm near}(i)} < {\rm Threshold}$, then $P \leftarrow {\rm Merge}(P, \bar{P}_i)$\\
14. end for

\bigskip

\revise{
Then, we explain how to obtain the objects' poses. 
If a segment of the currently captured depth image is similar to the previously captured one, we do not need to calculate the object's poses by using this segment of the depth image and can save the calculation time. 
}

\revise{
We first segment the merged depth image. For each segment, we calculate the distance between a point included in the segment and the surface of the object which pose is estimated during the previous picking trial. 
If the distance is less than the threshold, we use the result of pose estimation during the previous picking trial. 
On the other hand, if the distance is larger than the threshold, we newly estimate the pose of an object (Aldoma, 2012). 
}

\section{Picking Task Execution}

To pick up an object from the pile, we first capture the point cloud of the pile. 
Then, by using the method presented in Section \ref{sec:pose}, \revise{we determine sensor pose maximizing the visibility of the pile. 
Furthermore, by using the method presented in Section \ref{sec:est},} we consider estimating the objects' poses. 
For each object whose pose is estimated, we calculate candidates of grasping posture. 
For grasping postures where IK is solvable, we consider predicting whether or not a robot can successfully pick an object from the pile by using the discriminator constructed in Section \ref{sec:learn}. 
If it is predicted that a robot can successfully pick an object, the robot actually tries to pick the object. 
This section explains this pipeline of picking task execution. 

For a given object, a set of stable grasping configuration ${\cal G}$ with respect to the object coordinate system is calculated by using a grasp planner such as (Harada, 2008) 
in advance of a gripper actually grasps an object. This set can be defined as

\begin{equation}
{\cal G} = \left\{ (^o \bm{r}_i, ^o \bm{R}_i, \bm{\theta}_i, I_i), i=1,\cdots,d \right\}
\end{equation}

\noindent
where $^o\bm{r}_i$/$^o\bm{R}_i$ and $\bm{\theta}_i$ denote the position/orientation of the wrist with respect to the object coordinate system and the finger displacement vector, respectively. 
$I_i$ denote an index for evaluating the grasp stability such as (Harada, 2014). 

After estimating the objects' poses as 
$\bm{r}_{oj}$/$\bm{R}_{oj}$ $(j=1,\cdots,e)$, 
we obtain candidates of grasping configurations as

\begin{equation}
{\cal G}_c = \left\{ (\bm{r}_{ij}, \bm{R}_{ij}, \bm{\theta}_i, I_i), i=1,\cdots,d, j=1,\cdots,e \right\} \label{eq:set}
\end{equation}

\noindent
where $\bm{r}_{ij}=\bm{r}_{oj}+\bm{R}_{oj}^o\bm{r}_i$ and $\bm{R}_{ij}=\bm{R}_{oj}^o\bm{R}_i$. 
Here, solving IK (inverse kinematics) for all the elements of eq.(\ref{eq:set}) may take a lot of time especially when the database size $d$ is large. 
Hence, we consider splitting the candidate sets into $f$ subsets according to the grasp quality index 
$I_i$ as follows:

\begin{eqnarray}
{\cal G}_{c1} &=& \left\{ (\bm{r}_{ij}, \bm{R}_{ij}, \bm{\theta}_i, I_i), i=1,\cdots,d, j=1,\cdots,e | \right. \nonumber \\
              && \left. I_i > t_1 \right\} \nonumber \\
              &\vdots& \nonumber \\
{\cal G}_{ck} &=& \left\{ (\bm{r}_{ij}, \bm{R}_{ij}, \bm{\theta}_i, I_i), i=1,\cdots,d, j=1,\cdots,e | \right. \nonumber \\
              && \left. I_i \le t_{k-1}, I_i > t_k \right\} \nonumber \\
              &\vdots& \nonumber \\
{\cal G}_{cf} &=& \left\{ (\bm{r}_{ij}, \bm{R}_{ij}, \bm{\theta}_i, I_i), i=1,\cdots,d, j=1,\cdots,e | \right. \nonumber \\
              && \left. I_i \le t_{f-1} \right\}  
\end{eqnarray}

For grasping poses included in the $k$-th set ($k=1 \cdots f$), we calculate the IK and apply the random forest. 
The output of the random forest is the success rate $S_{ij}$. 

\begin{figure}
	\centering
	\includegraphics[width=9.cm]{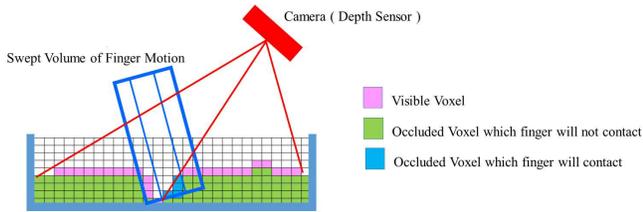}
	\caption{A heuristic rule to give penalty for occluded area included in the swept volume. \label{fig:occlusion}}
\end{figure}

Here, if a robot tries to pick an object from the pile by inserting the finger to an occluded area of a 3D depth sensor, the robot may fail in picking an object due to an unexpected contact between the finger and an object. 
We consider imposing penalty to a grasping posture where its swept volume of finger motion includes occluded area since we want to avoid unexpected contact between a finger and hidden neighboring objects. 
Overview of the method is shown in Fig. \ref{fig:occlusion}. As shown in this figure, we consider partitioning the storage area like Fig. \ref{fig:occupancy}. 
However, different from Fig \ref{fig:occupancy}, we differentiate two kinds of occluded grid cells drawn by green and blue colors. 
The occluded grid cells where the gripper will not contact during the approach phase are drawn by green color. 
On the other hand, the occluded grid cells where the gripper may contact during the approach phase are drawn by blue color. 
Since we want to avoid unexpected contact with neighboring object during the approach phase, we want to make the number of blue grid cells as small as possible. Let $n_{bi}$ be the number of blue grid cells, we subtract the output of the random forest 
$S_{ij}$ by $\alpha n_{bi}$. 

\section{Discussion}

\revise{
Our proposed method can be applied to a stack of objects with a relatively simple shape. If we consider applying our method to more complex shaped objects, we will encounter a problem which has not been addressed in this paper. If complex shaped objects are randomly stacked, two objects included in the pile may be tangled each other. If a robot tries to pick one of the tangled objects, a robot may fail in picking the target object since the neighboring object is not traversable. Or, even if the target object can be picked up, the neighboring object may be lifted at the same time. These unexpected cases are beyond the scope of this paper. This research deals with a stack of simple shaped objects where two objects included in the pile are not tangled each other. 
}

\section{Experiment}

We performed experiments on bin-picking. 
Overview of the robot system is shown in Fig. \ref{intro:RA-L2015}. 
We use the dual-arm manipulator HiroNX 
and its left hand to pick an object. 
Our HiroNX also has a 3D depth sensor (Xtion PRO) attached to the right wrist. 

\begin{figure}
	\centering
	\includegraphics[width=8cm]{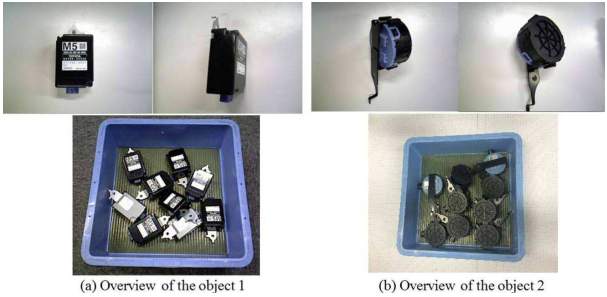}
	\caption{Objects used in experiment \label{fig:object}}
\end{figure}

As shown in Fig. \ref{fig:object}, we used two kinds of objects where one object has a cylinder-like shape and the other has a rectangular-like shape. 
For both cases, nine objects are randomly \revise{laid on the bottom surface of} a box. 
We put nine objects close to each other such that the finger contacts a neighboring object when picking the target one. 
\revise{To pick up an object from the pile, it would be easy for a robot to pick the topmost object since a finger will not contact a neighboring object during the approach to the target object. 
In our experiment, we consider a situation where the objects are laid on the bottom surface of a box. 
In this case, we can expect that a finger often contacts a neighboring object and it would be difficult for a robot to pick an object from the pile. }

\subsection{Discriminator Training}

To collect the training data, we used the visual recognition method explained in Section \ref{sec:est}. 
Then, candidates of grasping postures are obtained without considering the contact between the finger and neighboring objects. 
Among the candidates, we selected one with the highest grasp stability index (Harada, 2014). 
We performed a picking experiment and recorded as success if the target object is successfully lifted up from the pile. We collected 150 training data both for objects 1 and 2. 
Since we did not use any methods to avoid a finger contacting the neighboring objects, a robot failed in picking an object almost once per two trials. 
Among 150 training data, 71 were recorded as success for the object 1 and 
88 were recorded as success for the object 2. 

In Random Forest, the number of data included in the subset ${\cal L}_k$ is set as 70$\%$ of that of the set ${\cal L}$. 
Also, we used 25 dimensional feature vector by setting 
$b_y=b_z=5$, $w_y=w_z = 0.01${[m]}, $N=200$ and $t_k=5$. 

\begin{figure}
	\centering
	\includegraphics[width=8cm]{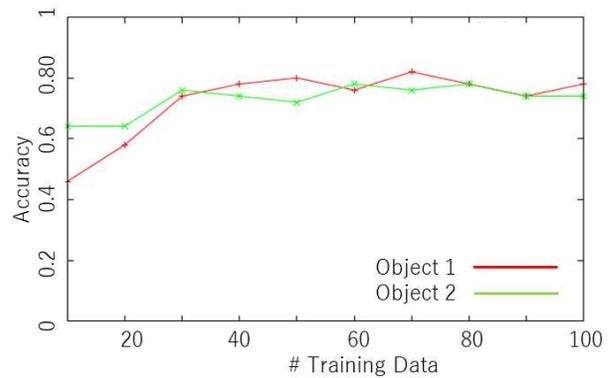}
	\caption{Estimation accuracy \label{fig:accuracy}}
\end{figure}

By using 150 training data, we performed the test sample estimation to check the accuracy of estimation. 
The result is shown in Fig. \ref{fig:accuracy}. For both objects 1 and 2, the accuracy is saturated when the number of training data is about 50. 
This result shows that, just for the purpose of predicting whether or not a robot successfully picks an object from the pile, the random forest with the small number of training data is enough. 
This result shows the effectiveness of using random forest for learning based randomized bin-picking since collecting a large number of training data by conducting picking experiment is burdensome. 

\subsection{Visual Recognition}

\begin{figure}
	\centering
	\includegraphics[width=9cm]{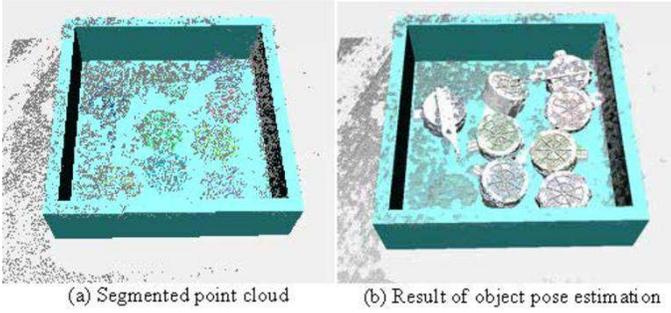}
	\caption{Estimation of objects' pose \label{fig:segmentation}}
\end{figure}

To check the validity of the iterative visual recognition method, we performed experiments on picking an object for three times. 
Fig. \ref{fig:segmentation} shows the result of segmentation and pose estimation of randomly stacked objects for the first picking trial where the pose of eight objects are estimated in this case. 

\begin{figure}
	\centering
	\includegraphics[width=9cm]{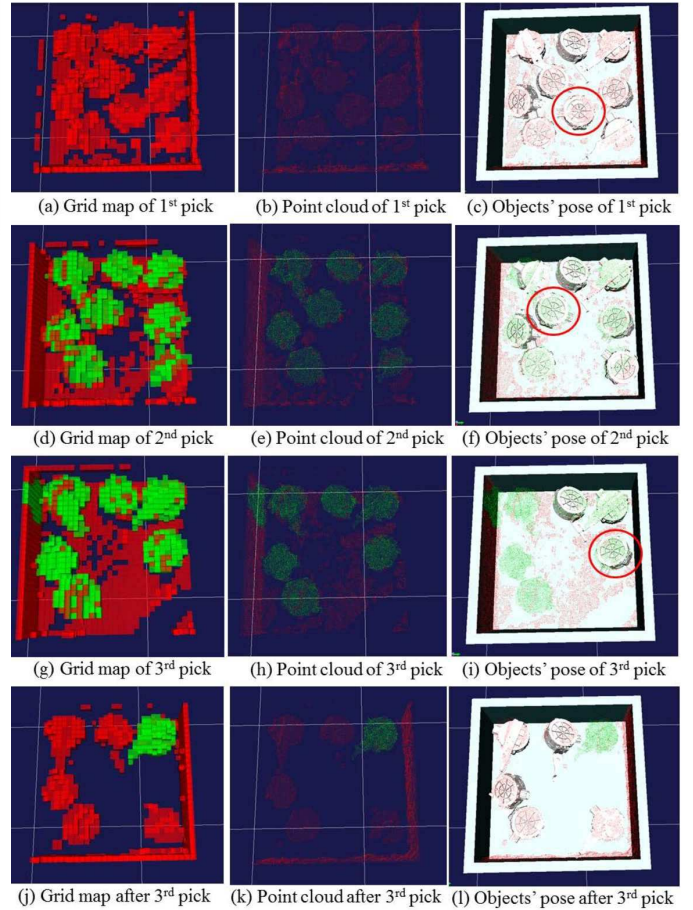}
	\caption{Grid cells of captured point cloud \label{fig:capture}}
\end{figure}

\begin{figure}
	\centering
	\includegraphics[width=8cm]{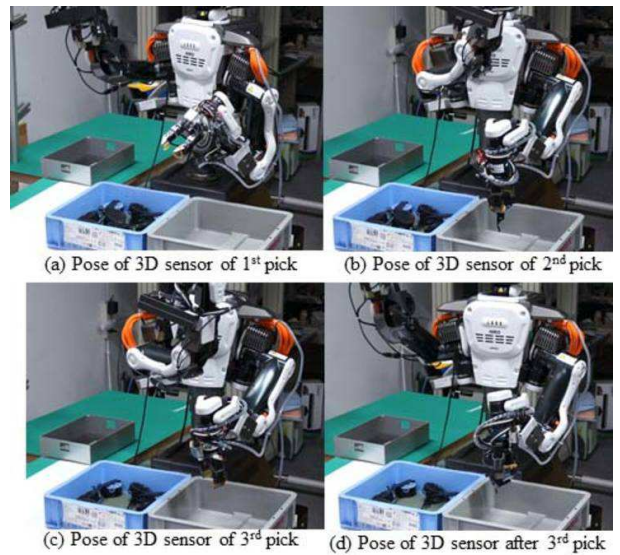}
	\caption{Pose of 3D sensor during a series of picking task \label{fig:camera}}
\end{figure}

Fig. \ref{fig:capture} shows the grid cells of captured point cloud during a series of picking tasks where the red cells include the newly captured point cloud while the green cells include the previously captured point cloud. 
During the 1st picking trial, poses of all nine objects are recognized. 
Here, the picked object is marked by the red circle. 
During the 2nd picking trial, point cloud belonging to all the objects are included in the green cells. This is because, since the finger did not contact the neighboring objects during the first picking trial, the configuration of objects did not change. 
We can see that object recognition is performed only for the object where red cells are included. 
In our current setting, since we limit the number of objects whose poses are detected to eight, some objects are failed in identification especially when two objects commonly belong to a single segment. 
Fig. \ref{fig:camera} shows the pose of 3D vision sensor during a series of picking task by using the dual-arm industrial manipulator HiroNX. We can see that the robot changes the view point for every picking trial to get more complete point cloud of randomly stacked objects. 

The calculation time for detecting the objects' poses is shown in 
Table \ref{tab:pose}. We used a PC with dual Xeon 3.33GHz CPUs. 
Also, we consider calculating the ICP in parallel. 
Regardless the number of threads used to calculate ICP, the calculation time increases with the growth of the detected objects. 

\begin{table*}
\centering
\caption{Calculation time used for detecting the objects' poses {[s]}\label{tab:pose}}
\begin{tabular}{|c|p{1cm}|p{1cm}|p{1cm}|p{1cm}|} \hline 
{\scriptsize Number of detected objects $\backslash$ Number of threads} & 1  &  2  & 4  & 8   \\ \hline
1 & 2.49 & 1.54 & 1.10 & 0.84 \\ \hline 
2 & 3.46 & 2.08 & 1.46 & 1.06 \\ \hline
3 & 4.78 & 2.83 & 1.88 & 1.31 \\ \hline
4 & 5.73 & 3.11 & 2.07 & 1.44 \\ \hline
5 & 6.24 & 3.66 & 2.40 & 1.63 \\ \hline
6 & 7.34 & 4.16 & 2.69 & 1.85 \\ \hline
7 & 7.98 & 4.57 & 2.94 & 2.01 \\ \hline
8 & 8.53 & 4.83 & 3.13 & 2.10 \\ \hline
9 & 9.16 & 5.14 & 3.33 & 2.22 \\ \hline
\end{tabular}
\end{table*}

\subsection{Picking Experiment}

We performed picking experiments to see the effectiveness of our proposed method. 
In our experiment, if the discriminator predicts that a robot will fail in picking an object, we captured the point cloud again from different viewpoint and merged it to the existing point cloud by using the method explained in Section \ref{sec:pose}. 
Then, we recognized the objects' poses again by using the method explained in Section \ref{sec:est}. 
We iterated this procedure for maximum three times until the discriminator predicts the success. Since we subtracted the output of Random Forest algorithm according to the amount of occluded area included in the swept volume, the discriminator tends to predict the failure if the point cloud includes a lot of occluded area. 
However, if we merge the point cloud from different viewpoint, the discriminator tends to \revise{predict} the success according to the decrease of occluded area. 

The results of this experiment is shown in Tables \ref{tab:rand1} and \ref{tab:rand2}. 
We can see that the F-value is high enough in both cases. Among 50 cases predicted as success for the object 1, 27 cases were initially predicted as failure.  
On the other hand, among 50 cases predicted as success for the object 2, 25 cases were initially predicted as failure. This result shows that, since we give penalty for a part of point cloud including occlusion, a robot did not initially find a grasping pose. 
On the other hand, after capturing point cloud for three times from different point of view, we can obtain more complete point cloud and a robot can successfully find a grasping pose. 

Then, we checked a finger actually contacts a neighboring object. 
Among 40 cases that is predicted as success and actually succeeded in picking the object 1, the finger contacted neighboring objects for 13 cases. 
On the other hand, among 10 cases predicted as success but failed in picking the object 1, the finger contacted neighboring objects for 7 cases. 
Also, among 46 cases predicted as success and succeeded in picking the object 2, the finger contacted neighboring objects for 17 cases. 
On the other hand, among 4 cases predicted as success but failed in picking the object 2, the finger contacted neighboring objects for 2 cases. 

Throughout the experiment, a robot sometimes failed in picking an object even if it was predicted to successfully pick an object. This is mainly due to the error \revise{of} segmentation and pose identification. 
Also, the precision of the experiment using the object 1 is lower than the precision of the experiment using the object 2. This is because, since the object 1 has a rectangular-like shape, the effect of identification error of its orientation to the result of picking becomes larger especially when a robot tries to pick the rectangular near its edge. 

\begin{table}
\caption{Results of picking experiment of the object 1 where Precision:0.8, Recall:0.98, F-value:0.88\label{tab:rand1}}
\begin{tabular}{|l|c|c|} \hline 
 & {\scriptsize Picking succeeded} & {\scriptsize Picking failed} \\ \hline
{\scriptsize Predicted as succeess} & 40 & 10 \\ \hline 
{\scriptsize Predicted as failure} & 1  & 5 \\ \hline
\end{tabular}
\end{table}

\begin{table}
\caption{Results of picking experiment of the object 2 where Precision:0.92, Recall:1.0, F-value:0.96\label{tab:rand2}}
\begin{tabular}{|l|c|c|} \hline 
 & {\scriptsize Picking succeeded} & {\scriptsize Picking failed} \\ \hline
{\scriptsize Identified as \revise{success}} & 46 & 4 \\ \hline 
{\scriptsize Identified as failure} & 0  & 1 \\ \hline
\end{tabular}
\end{table}

\begin{figure}
	\centering
	\includegraphics[width=6.5cm]{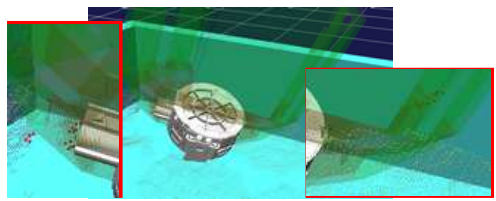}
	\vspace{-1mm}\\
	{\scriptsize (a) Swept volume of finger motion}
	\vspace{1mm}\\
	\includegraphics[width=5cm]{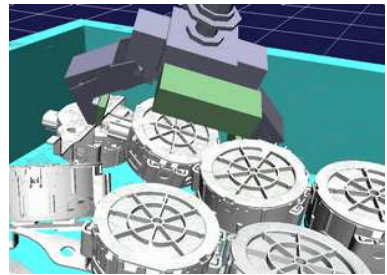}
	\vspace{-1mm}\\
	{\scriptsize (b) Calculated grasping posture}
	\vspace{-2mm}
	\caption{Result of grasping posture planning \label{fig:grasp}}
\end{figure}

\begin{figure}
	\centering
	\includegraphics[width=7cm]{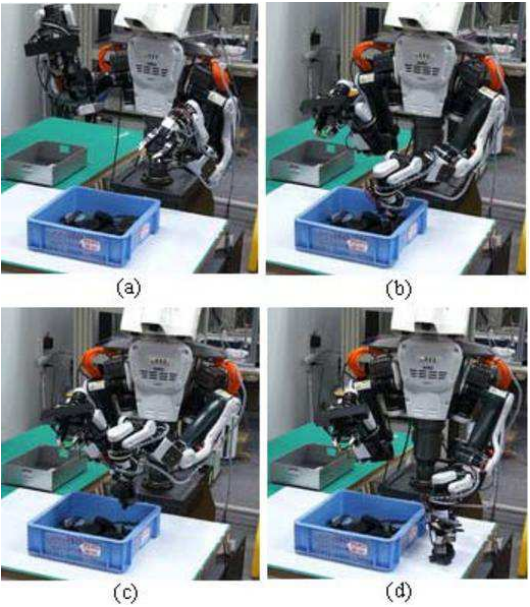}
	\caption{Overview of picking experiment \label{fig:experiment}}
        \vspace{-0.4cm}
\end{figure}

Figs. \ref{fig:grasp} and \ref{fig:experiment} show a series of experimental result. 
For the object, we prepared the grasping configuration database where its size is $d=172$. 
We split the candidates of grasping configuration into three subsets ($f=3$) where we set $t_1$ and $t_2$ such that the size of each subset becomes as same as possible. 
We could find a feasible grasping configuration from ${\cal G}_{c1}$. Among $d  e / f \simeq 453$ candidates, 98 were IK solvable. Then, 50 out of 98 were identified to be a successful case of picking by using the random forest. 
Fig. \ref{fig:grasp} shows the selected grasping configuration and its finger swept volume. 
Here, in Fig. \ref{fig:grasp}(a), the red dot shows the point cloud included in the finger swept volume. 
Here, to cope with the sensor noise, we assumed a small margin (0.002{[m]}) to the size of the finger swept volume. 
Hence, in the figure, we can find some red dots out of the finger swept volume. Finally, Fig. \ref{fig:experiment} shows a series of experiment snapshot. In the experiment, although the finger contacts a neighboring object, the robot can successfully perform the picking task. 

\section{Conclusions}

In this paper, we proposed a learning based approach on randomized bin-picking. 
By using the distribution of the point cloud, the discriminator predicts whether or not the picking will be successfully performed even if a finger contacts a neighboring object. 
We also explained the view planning method to get more complete information on the randomly stacked objects. 
Since randomized bin-picking usually estimates the pose of a number of objects, we relaxed the computational cost of the object pose detection by using the visual information on randomly stacked objects captured during the current picking task together with the visual information captured during the previous picking tasks. 
The effectiveness of the proposed approach was confirmed by a series of experimental results. 

The followings are some of the remaining problems: 
First, performance of picking may further increase if we use time-series visual information to train the discriminator. 
Second, extension of our proposed algorithm to more general multi-fingered hand is also considered to be our future research topic. 
\revise{Thirdly, we will more explicitly evaluate the occluded area and show that 3D visual image of the pile with less occluded area is effective in predicting the success/failure of a pick. We will also evaluate the number of image required to detect the objects' poses with sufficient precision.}

\end{document}